\documentclass[10pt,twocolumn,letterpaper]{article}

\usepackage[pagenumbers]{wacv}            

\usepackage{graphicx}
\usepackage{booktabs}
\usepackage{amsmath}
\usepackage{amssymb}
\usepackage{makecell}
\usepackage{multirow}
\usepackage{xcolor}

\newcommand{\dofull}{\ensuremath{d_0/\text{full}}}        
\newcommand{\R}[1]{\ensuremath{\text{R@}#1}}

\definecolor{wacvblue}{rgb}{0.21,0.49,0.74}
\usepackage[breaklinks,colorlinks,allcolors=wacvblue]{hyperref}

\def\wacvPaperID{913}
\def\confName{WACV}
\def\confYear{2027}

\title{Object-Only Is Not Context-Free:\\A Counterfactual Diagnostic
for Context Removal in Object Retrieval}

\author{Boaz Meivar \qquad Shai Avidan\\
Tel Aviv University\\
{\tt\small boazmeivar@mail.tau.ac.il \quad avidan@eng.tau.ac.il}}

\begin{document}
\maketitle

\begin{abstract}
Object-centric retrieval pipelines often mask the background before
pooling backbone features, aiming for a descriptor that reflects the
object rather than its scene. Whether this truly yields an object-only
descriptor, however, is rarely tested. We introduce the \emph{Keyhole
Aperture Probe} (KAP), a counterfactual diagnostic that repurposes the
ground-truth masks of existing panoptic segmentation datasets
(COCO, ADE20K) for retrieval: it holds the pooled object region fixed
and varies only the pixels outside it, summarizing a backbone's context
dependence in a single number. We find that the pooled object region
has already integrated its surround. Replacing the context, with a
neutral fill or even a different real scene, changes the descriptor,
and when retrieval fails under a blank fill, the descriptor matches
plain-background images rather than its own object. Object-only is not
context-free. Across ten modern backbones, object-only descriptors
retain only about a quarter to two-thirds of their full-image accuracy,
and the effect is not unique to Transformer self-attention: a
convolutional backbone shows it too. The dependence is also
reducible: a lightweight adapter that encourages the descriptor to stay
stable as context is removed lowers it and \emph{improves} full-image
instance retrieval on INSTRE, transferring to ILIAS with accuracy
improved or preserved. A matched contrastive-fine-tuning control does
not reproduce those gains. We \href{https://github.com/bmeivar/kap}{release} KAP as a
simple, backbone-agnostic evaluation protocol to measure and reduce
this dependence before deployment.
\end{abstract}

\section{Introduction}
\label{sec:intro}

\begin{figure*}[t]
  \centering
  \includegraphics[width=0.92\linewidth]{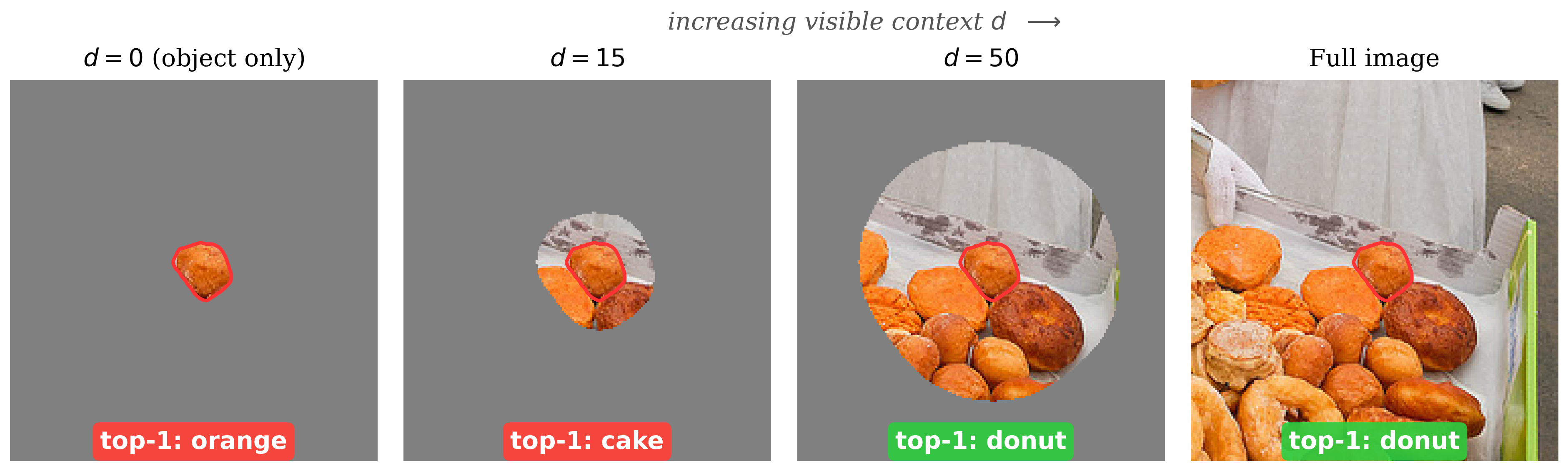}
  \caption{\textbf{How much context does an object descriptor need?}
  A COCO donut query under an expanding visible aperture (gray fill),
  with DINOv2-G/14 top-1 retrievals. At $d{=}0$ (mask only) the top-1
  is wrong (\emph{orange}); as the aperture opens the match flips
  through \emph{cake} to the correct \emph{donut}. KAP reduces this
  dependence to one scalar; across ten backbones it varies widely and
  is \emph{not exclusive to ViT self-attention}.}
  \label{fig:teaser}
\end{figure*}

Object-level retrieval is routinely framed as an instance-isolation
problem: obtain a mask, pool features inside it, and treat everything
outside as nuisance. Yet modern foundation encoders are trained on
full images and their co-occurrence
structure~\cite{oquab2023dinov2,zhai2023siglip}, so an ``object
descriptor'' pooled from a masked region likely summarizes object
\emph{plus} context. Some leakage is expected, since attention and large
receptive fields mix every token with its surround; what is missing is
a measurement of how much, for which backbones, and through what
mechanism. An object stripped to its mask can even
retrieve the wrong category outright (\cref{fig:teaser}). Context is
informative for identity, location, and
scale~\cite{torralba2003context}, but reliance on it induces shortcut
behavior and background bias that fail under
shift~\cite{geirhos2020shortcut,xiao2021noise}; for retrieval there is
little controlled, at-scale evidence on which effect dominates or on
what makes a backbone susceptible. Masking, moreover, does not so much
remove context as replace it with a neutral fill, and we show
(\cref{sec:stratified}) that the encoder folds this fill back into the
``object-only'' descriptor.

We address this with a \emph{counterfactual} measurement built on
panoptic segmentation
datasets~\cite{lin2014coco,kirillov2019panoptic,zhou2017ade20k}, whose
pixel-precise masks let us delineate an object exactly and intervene
only on its surroundings: per query we hold fixed the object pixels,
pooling region, database, and labels, and change \emph{only} the
pixels outside the mask, sweeping an expanding aperture
($d{=}0,5,\dots,50$\,px and full image), gray-filling outside it,
while always pooling over the original undilated mask. This decouples
\emph{what the encoder sees} from \emph{what we pool}. We summarize
each backbone with one scalar, \dofull{}: the fraction of full-image
retrieval accuracy retained when only object pixels are visible. We
call this protocol the Keyhole Aperture Probe (KAP); it reads out
through nearest-neighbor retrieval, and the same ranking appears under
a trained classifier (\cref{sec:validity}).

We apply KAP across ten diverse backbones and find that all show
substantial context reliance. When a descriptor breaks, it drifts toward
plain-background images rather than its own object; the effect is not
specific to Transformer attention, and sharpest for small objects.

Why does this matter, if the full image is usually available? The
sweep is a measurement instrument, and the cost it exposes bites in
two deployment regimes. \emph{(i)}~Masking the query to obtain
``object-specific'' features (a masked-crop-then-encode step used
ahead of frozen backbones in open-vocabulary
pipelines~\cite{ding2022zegformer,xu2022simplebaseline,liang2023ovseg})
implicitly assumes a matching masked gallery. Against a \emph{natural}
gallery the masked query is distribution-shifted and loses.
\emph{(ii)}~Even with full images on both sides, a context-reliant
descriptor matches on background, so the same object in a new scene
retrieves poorly. The reliance is also reducible: an
aperture-conditioned training signal yields features that improve
full-image instance retrieval on INSTRE~\cite{wang2015instre} ($+3.6$
to $+11.9$ mAP) and transfer to
ILIAS~\cite{kordopatiszilos2025ilias}, with a matched control ruling
out generic contrastive fine-tuning as the cause. The rest of the
paper measures this dependence, validates it as a diagnostic, traces
its mechanism, and reduces it.

\paragraph{Contributions.}
\begin{itemize}
\itemsep0.2em
\item \textbf{A diagnostic repurposed from segmentation data.} We turn
panoptic segmentation masks into KAP, a counterfactual evaluation
protocol over existing datasets (no new data collection required) that fixes
the pooled object region and varies only the surrounding context,
reducing a backbone's reliance on context to a single number. We show
this number reflects the representation, not our measurement choices: it
is stable to the fill operator, soft edges, imperfect masks, and even a
non-retrieval (classifier) readout (\cref{sec:validity}).
\item \textbf{A mechanism for why removal hurts.} Object-only is not
context-free: when a query breaks, its descriptor becomes closest to
plain-background images rather than to its own object, because the
encoder folds the blank fill into the pooled features
(\cref{sec:stratified}).
\item \textbf{A controlled finding about what drives it.} Context
reliance is not unique to Transformer self-attention: a convolutional
network shows it too, in a controlled architecture comparison at matched
scale, data, and resolution. The effect is far stronger for small
objects (\cref{sec:breadth,sec:stratified}).
\item \textbf{An actionable fix.} A lightweight aperture-conditioned
adapter reduces context reliance and \emph{improves} full-image instance
retrieval on INSTRE under predicted masks, transferring to ILIAS
(accuracy improved or preserved) under a different mask source; a
matched contrastive-fine-tuning control shows the gain is
specific to the aperture curriculum (\cref{sec:instre}).
\end{itemize}

\section{Related Work}
\label{sec:related}

\paragraph{Retrieval representations and pooling.}
Deep image retrieval has progressed from compact global descriptors to
hybrid local--global systems, and the pooling operator strongly shapes
which regions a descriptor emphasizes. Generalized-mean pooling
(GeM)~\cite{radenovic2018gem} became a standard component of strong
global descriptors, while attention-based local-feature selection
(DELF)~\cite{noh2017delf} and unified local--global
models~\cite{cao2020delg} target robustness to clutter and partial
visibility; region aggregation over CNN activations is a classic
precursor~\cite{tolias2016rmac}. Mask pooling over a frozen backbone's
feature map is itself common practice, revisited recently for
region-level recognition~\cite{shlapentokh2024region} and used to turn
frozen vision--language features into open-vocabulary
segmenters~\cite{yu2023fcclip}, including retrieval-augmented
few-shot ones that pool frozen VLM features inside SAM
masks~\cite{aravanis2026rns}. A recurring finding in that literature
is that even an accurate mask does not guarantee an accurate
readout: Retrieve and Segment reports residual ambiguity despite
well-aligned masks~\cite{aravanis2026rns}, and mask quality alone
leaves a gap that downstream mask-refinement stages are built to
close~\cite{liang2023ovseg,li2025maskadapter}.
We do not propose a new descriptor. We evaluate the
\emph{object-descriptor abstraction}, masked pooling over a backbone
feature map, under controlled background interventions, asking why
that gap exists at the representation level rather than closing it
with a better mask.

\paragraph{Foundation backbones and pretraining paradigms.}
Self-supervised~\cite{caron2021dino,oquab2023dinov2},
vision--language contrastive~\cite{zhai2023siglip}, masked-image-modeling
~\cite{he2022mae,fang2024eva02}, and supervised
~\cite{liu2022convnext} encoders are increasingly used off the shelf for
retrieval. Because all are trained on full images, their
representations may implicitly encode scene regularities and
object--context co-occurrence. We sample backbones across these
families, at comparable scale and with matched-condition controls
(\cref{sec:breadth}), to ask
whether context dependence tracks the architecture, the training
objective, or neither, rather than to crown a single best model.

\paragraph{Context, shortcuts, and counterfactual analysis.}
Classic work established that scene statistics prime recognition and
carry predictive information about objects~\cite{torralba2003context},
and the cognitive-science lineage treats context as integral to
recognition~\cite{oliva2007role}, most useful exactly when local
evidence is weak, as for small or degraded objects. Our size-graded
result (\cref{sec:stratified}) is the representation-level counterpart,
with one difference: a learned descriptor imports context even when the
task asks it not to. KAP asks the analogous question of a
learned representation's pooled features, while shortcut
learning~\cite{geirhos2020shortcut} and background-bias
studies~\cite{xiao2021noise} show such cues generalize poorly.
Methodologically, our intervention is a counterfactual edit in the
spirit of perturbation-based explanation~\cite{fong2017meaningful},
applied at retrieval scale. Panoptic
annotations~\cite{lin2014coco,kirillov2019panoptic,zhou2017ade20k}
serve as an \emph{oracle measurement instrument}, not a deployment
requirement: the conclusions survive predicted
masks~\cite{sam3} (\cref{sec:validity,sec:instre}).

\paragraph{Scope relative to object-centric representation learning.}
This paper studies the practice of hard segment-then-pool masking as
deployed in retrieval systems~\cite{korfhage2024searchanything} and
assumed by instance-retrieval benchmarks that construct context-free
queries~\cite{kordopatiszilos2025ilias} (\cref{sec:instre}), not the
design of a new pooling mechanism. Slot attention~\cite{locatello2020slotattention} and its
follow-ups~\cite{seitzer2023dinosaur,jia2023boqsa} pursue a soft,
object-centric alternative whose slots are \emph{discovered} rather
than tied to a fixed external mask and are optimized to bind to one
object each, exactly the failure mode a hard mask cannot prevent. KAP
applies to slot read-outs directly: probing
FT-DINOSAUR~\cite{didolkar2024ftdinosaur} on COCO (each object
represented by the slot whose decoder mask best covers it) against
mask-pooling the \emph{same} encoder's features, the slot descriptor
still loses over half its accuracy under context removal (\dofull{}
$42.5\%$ vs.\ $33.0\%$ mask-pooled, val2017). Its higher retention
reflects a lower full-image ceiling ($-11.5$ R@1, 95\% CI
$[-13.2,-9.8]$), not better object-only accuracy (tied at $d_0$): slot
binding reduces how much context the descriptor \emph{uses}, but does
not yield a context-free representation. Slot attention is therefore
complementary: one candidate, partial mechanism for closing the gap
KAP quantifies.

\section{The Keyhole Aperture Probe}
\label{sec:method}

\paragraph{Setup and notation.}
Given an image $I$ and a ground-truth object mask $M_O$, a backbone
$f$ produces a dense feature map $f(I)$, from which we form an object
descriptor by mask pooling, $v(I,M_O)=\mathrm{norm}\!\big(\mathrm{pool}_{M_O}
f(I)\big)$, with mean pooling and $\ell_2$ normalization unless noted.
Throughout, we study this \emph{mask-pooled} object descriptor, which
is common in region-centric
pipelines~\cite{shlapentokh2024region,yu2023fcclip}. We keep two
operations distinct: \emph{input masking} replaces pixels before
encoding, while \emph{mask pooling} restricts which tokens are averaged
after encoding. Region-based retrieval pipelines use one or both;
every KAP condition uses both, with the input mask at the aperture and
the pooling mask always at the object.
Retrieval is category-level object--object: a query descriptor is
ranked against a database of object descriptors and scored by
recall@$K$ (R@$K$, the fraction of queries with a correct match in the
top $K$), mean average precision (mAP), and mean reciprocal rank (MRR),
with same-image matches excluded so a query is
never scored against objects from its own image (image-disjoint
evaluation). Masks are used only to \emph{define} the query
intervention and pooling support; the database is encoded once from
natural images. This asymmetry is deliberate: the deployed gallery is
natural and indexed once, and masking both sides would answer a
different question, confounding the effect we isolate with the
mask-only failure mode of \cref{fig:teaser}.

\begin{figure*}[t]
  \centering
  \includegraphics[width=0.78\linewidth]{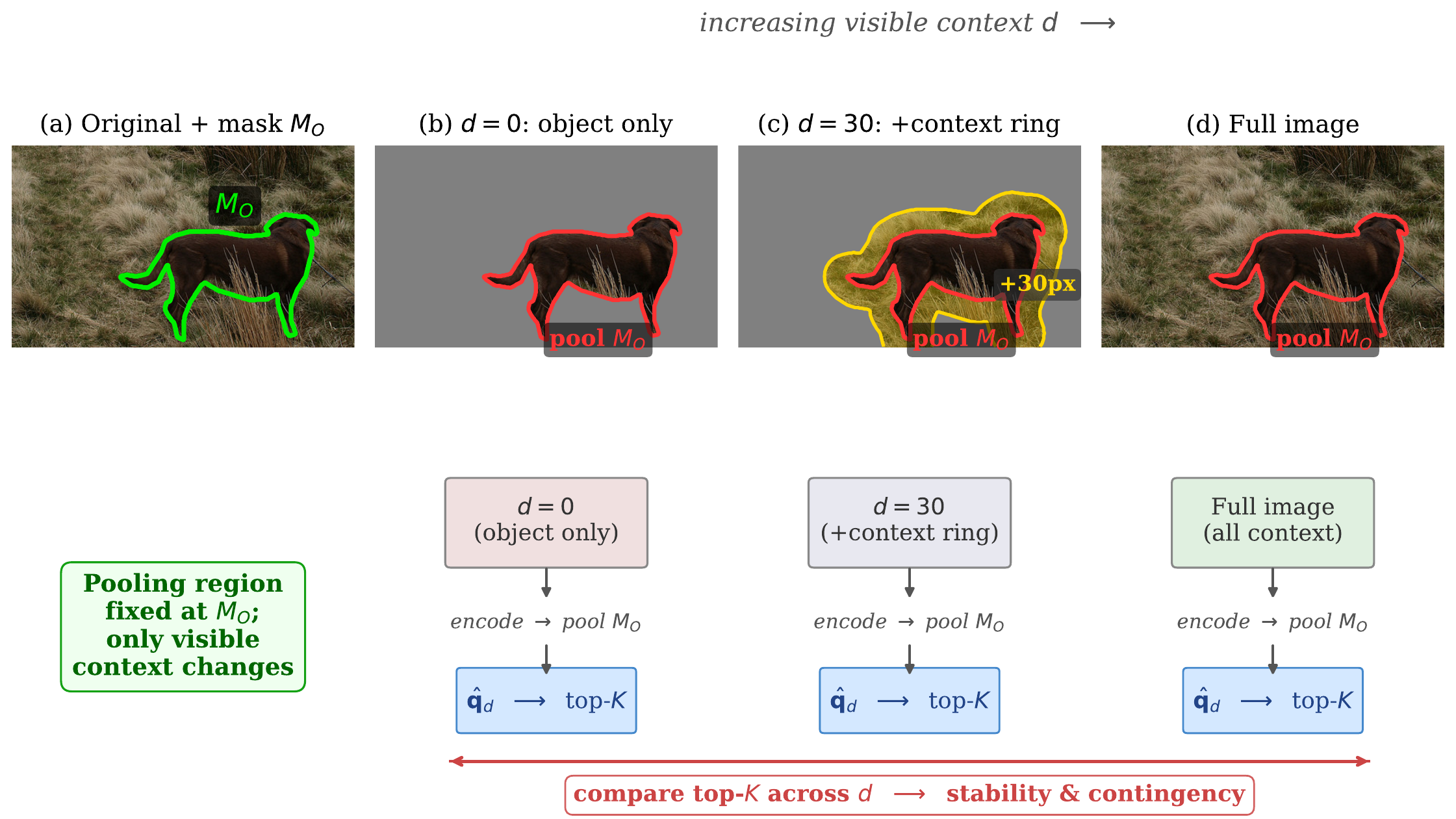}
  \caption{\textbf{The Keyhole Aperture Probe.} The visible aperture
  $M_d$ expands with $d$ (outside-mask pixels replaced by a neutral
  fill), but the descriptor is \emph{always} pooled over the original
  object mask $M_O$, so $d$ changes only the context the encoder
  integrates, never the read-out tokens. Comparing the top-$K$ neighbor
  set across $d$ isolates the effect of visible context on the
  object-only descriptor.}
  \label{fig:protocol}
\end{figure*}

\paragraph{The aperture intervention.}
The probe varies how much context the encoder sees while holding the
pooled support fixed (\cref{fig:protocol}). For a dilation radius $d$ we define the visible
aperture $M_d=\mathrm{dilate}(M_O,d)$ and construct an intervened image
\begin{equation}
\tilde I_d(p) =
\begin{cases}
I(p), & p \in M_d,\\
\phi(p), & p \notin M_d,
\end{cases}
\end{equation}
where $\phi$ is a fill operator (neutral gray by default; zero,
blur, and alpha-feather are studied in \cref{sec:validity}).
The descriptor is always pooled over the \emph{original} mask,
$v(\tilde I_d, M_O)$, so $d$ changes only the context the encoder
integrates, never the set of tokens we read out. Sweeping
$d\in\{0,5,10,15,30,50\}$\,px and the full image traces a recovery
curve from ``object only'' ($d{=}0$) to ``full context'' (full).

\paragraph{A single context-dependence scalar.}
We summarize each backbone by its \emph{object-only retention}
\begin{equation}
\dofull \;=\; \frac{\R{1}(d{=}0)}{\R{1}(\text{full})},
\end{equation}
the fraction of full-image accuracy retained when only object pixels
are visible. Higher \dofull{} means the backbone retains more retrieval
accuracy without context. Concretely, \dofull{} is the retrieval cost of
removing context from the \emph{query} while the gallery stays natural,
the deployed masked-query setting, rather than a claim that
the backbone cannot represent the isolated object in principle. The two
aperture extremes bracket the counterfactual precisely: \emph{for the
same object, same pooling region, same database and labels, what is
retrieval accuracy with context removed versus restored?} Because only
outside-mask pixels differ between the two conditions, any gap is
attributable to content outside the object, not to the object, the
pooling support, or spatial position (\cref{sec:validity}).

\paragraph{Data and sampling.}
We run KAP on COCO-Panoptic~\cite{lin2014coco,kirillov2019panoptic}
\texttt{train2017} (the protocol of the breadth and validity
experiments; \cref{sec:instre} adds INSTRE): a
database of ${\sim}845$K ``thing'' instances and a class-balanced query
set ($40$/category, ${\sim}3.2$K queries); a cross-dataset replication
on ADE20K~\cite{zhou2017ade20k} is reported in the supplementary
material. Oracle
masks make the intervention exact; \cref{sec:validity,sec:instre} show
the diagnostic is robust to imperfect and \emph{predicted} masks, so
the oracle is a measurement convenience rather than a deployment
assumption.

\section{Results}
\label{sec:results}

\subsection{Context Dependence Across Backbones}
\label{sec:breadth}

\begin{table}[t]
\centering
\small
\setlength{\tabcolsep}{5pt}
\begin{tabular*}{\columnwidth}{@{\extracolsep{\fill}}llccccc@{}}
\toprule
Backbone & Pretr. & $d_0$ & full & \dofull & {\scriptsize 95\% CI} & AUC \\
\midrule
DINOv2-G/14      & SSL & 59.3 & 88.2 & 67 & {\scriptsize[65,\,69]} & 74 \\
ConvNeXt-L       & Sup & 45.0 & 68.2 & 66 & {\scriptsize[64,\,68]} & 50 \\
DINOv3-L         & SSL & 47.8 & 79.8 & 60 & {\scriptsize[58,\,62]} & 58 \\
ViT-L            & Sup & 44.2 & 75.1 & 59 & {\scriptsize[57,\,61]} & 54 \\
MAE ViT-L        & MIM & 13.7 & 24.1 & 57\rlap{$^\dagger$} & {\scriptsize[52,\,61]} & 16 \\
SigLIP2-512      & V-L & 47.1 & 85.8 & 55 & {\scriptsize[53,\,57]} & 62 \\
SigLIP2-L        & V-L & 40.7 & 81.2 & 50 & {\scriptsize[48,\,52]} & 53 \\
EVA-02-CLIP      & V-L & 37.0 & 76.7 & 48 & {\scriptsize[46,\,50]} & 44 \\
DINO ViT-S/8     & SSL & 18.0 & 56.0 & 32 & {\scriptsize[30,\,34]} & 27 \\
EVA-02-MIM       & MIM & 18.5 & 66.3 & 28 & {\scriptsize[26,\,30]} & 24 \\
\bottomrule
\end{tabular*}
\caption{\textbf{KAP across ten backbones} (COCO \texttt{train2017},
gray fill). Object-only ($d_0$) and full-image R@1 (\%), object-only
retention \dofull{} (\%) with paired query-bootstrap 95\% CIs
($10^4$ resamples), and AUC; sorted by \dofull{} (higher is more
context-resistant). Pretr.: pretraining paradigm, one of
self-supervised (SSL), vision--language (V-L), masked-image-modeling
(MIM), supervised (Sup).
AUC is the area under the recovery curve (the trapezoidal mean R@1
over the sweep; per-aperture curves in the supplementary material),
ranking the panel by absolute quality.
$^\dagger$MAE's \dofull{} ratio is unreliable: its full-image accuracy
is the lowest in the panel ($24.1$), inflating the ratio, and AUC ranks
it last.}
\label{tab:breadth}
\end{table}

We run KAP on ten backbones chosen to populate four pretraining
paradigms at comparable scale (\cref{tab:breadth}): self-supervised
(DINOv2-G/14~\cite{oquab2023dinov2},
DINOv3-L~\cite{simeoni2025dinov3}, DINO ViT-S/8~\cite{caron2021dino}),
vision--language contrastive
(SigLIP2~\cite{tschannen2025siglip2} at two resolutions,
EVA-02-CLIP~\cite{fang2024eva02}), masked-image-modeling
(MAE ViT-L, EVA-02-MIM), and supervised classification (ConvNeXt-L,
ViT-L), all under the identical protocol and database. Exact
checkpoints, input resolutions, and feature grids are in the
supplementary material.

\paragraph{Not unique to self-attention.}
A convolutional backbone shows the effect despite having no global
self-attention: ConvNeXt-L retains only $66\%$ of its full-image
accuracy from object pixels alone. Context dependence therefore does
not \emph{require} self-attention. If anything, the matched
supervised ViT-L is slightly \emph{more} affected ($59\%$, $+7$\,pp
gap; paired bootstrap 95\% CI $[+4.7,+9.3]$, $p<0.001$). The common
cause is \emph{spatial context integration}: by its deep layers a
ConvNeXt's effective receptive field spans much of the image, so an
object-region feature has already aggregated its surround. Global
attention and large receptive fields are two routes to the same leak.
This predicts the effect should grow as the object occupies less of
the receptive field, borne out by the size-grading of
\cref{sec:stratified}. The gap \emph{replicates under control}: with
ViT-B/16 and ConvNeXt-B matched at Base scale, $224$\,px, ImageNet-1k,
pretrain-only, the supervised ConvNeXt again retains more ($56.2$ vs.\
$48.9$). We restrict the claim to architecture, not pretraining
objective: the matched masked-image-modeling models sit at near-floor
accuracy (R@1 $22$, $13$), leaving the objective axis unmeasurable.

\paragraph{Recovery curves.}
Recovery is gradual: no backbone regains full accuracy until
substantial context is restored (per-aperture curves in the
supplementary material). AUC (the mean R@1 over the sweep,
\cref{tab:breadth}) orders the panel by overall retrieval quality,
placing DINOv2-G first and MAE last, where the \dofull{} ratio alone
ranks MAE mid-panel despite its near-floor accuracy. The two answer
different questions (context-resistance vs.\ absolute quality) and we
report both.

\paragraph{Cross-dataset replication (ADE20K).}
The findings are not specific to COCO: re-running the full ten-backbone
panel on ADE20K~\cite{zhou2017ade20k} (scene-parsing instances) under
the identical protocol preserves the backbone ranking (Spearman
$\rho{=}0.94$, $p<10^{-3}$ against COCO \dofull{}) and reproduces the
architecture dissociation (ConvNeXt-L again exceeds ViT-L, $+8$\,pp);
the full per-backbone table is in the supplementary material.

\subsection{What Context Removal Does}
\label{sec:stratified}

Beyond the single scalar, KAP shows \emph{how} context removal
reshapes retrieval.

\paragraph{Removal breaks far more than it fixes.}
Treating each query as broken or fixed by context removal relative to
the full-image condition reveals a stark asymmetry
(supplementary Fig.~2c): across all ten backbones removing context
\emph{breaks} $4$--$52\times$ more queries than it \emph{fixes}
($16{:}1$ for DINOv2-G, $30{:}1$ for SigLIP2-L, up to $52{:}1$ for
EVA-02-MIM). This is not isolated noise: the top-$10$ neighbor set
drops to barely $10\%$ overlap with the full-context ranking at $d{=}0$
(supplementary Fig.~2b), so context removal does not merely lower
accuracy. It reshapes which objects look similar. With the object
alone a bench retrieves an \emph{airplane}, a frisbee a \emph{cake},
scissors a \emph{cell phone}, and a toaster a \emph{remote}: the
isolated shape is read as a different object. A thin context ring barely
helps: $984$ queries still flip at
$d_5$, nearly as many as the $987$ at $d_0$. For the backbones tested,
``mask everything but the object'' is more often harmful than
helpful.

\begin{table}[t]
\centering\small
\setlength{\tabcolsep}{6pt}
\begin{tabular*}{\columnwidth}{@{\extracolsep{\fill}}lcccc@{}}
\toprule
& \multicolumn{3}{c}{\dofull{} by area tercile} & \\
\cmidrule(lr){2-4}
Backbone & small & mid & large & $\rho$ \\
\midrule
DINOv2-G/14    & 30 & 67 & 98 & 0.59 \\
ConvNeXt-L     & 24 & 61 & 92 & 0.58 \\
DINOv3-L       & 20 & 51 & 93 & 0.63 \\
ViT-L          & 22 & 49 & 90 & 0.59 \\
MAE ViT-L      & 33 & 42 & 72 & 0.45 \\
SigLIP2-512    & 20 & 48 & 88 & 0.58 \\
SigLIP2-L      & 18 & 37 & 84 & 0.59 \\
EVA-02-CLIP  & 15 & 34 & 81 & 0.61 \\
DINO ViT-S/8   & 18 & 20 & 47 & 0.36 \\
EVA-02-MIM   &  4 &  9 & 54 & 0.59 \\
\midrule
Mean  & 20 & 42 & 80 & 0.56 \\
\bottomrule
\end{tabular*}
\caption{\textbf{Context dependence is graded by object size: retention
climbs from $20$ to $80$ from small to large objects} (mean over the ten
backbones, bottom row). Per-backbone object-only retention \dofull{} (\%)
by query-object area tercile, and Spearman $\rho$ between object area and
survival of context removal (positive for all ten). Small objects
collapse without context; large objects retrieve nearly as well.}
\label{tab:sizestrat}
\end{table}

\paragraph{Small objects are the most context-dependent.}
The cost is strongly graded by object size (\cref{tab:sizestrat}). Within
each backbone, per-query survival of context removal correlates with
object area (mean Spearman $\rho{=}0.56$ over the ten backbones, positive
for every one). Stratifying by area, object-only retention rises from
$4$--$33\%$ for the smallest area tercile to $47$--$98\%$ for the
largest: large objects retrieve nearly as well without context, while
small instances collapse. This is the regime where the
``mask-everything-else'' heuristic is most damaging and, because small
objects occupy few patches, where added context, not just added pixels,
does the most work.

\paragraph{Why removal breaks: a plain-background crossover.}
What does a context-removed descriptor become similar to? In the
cosine metric retrieval uses, we compare each query's similarity to
its own-category centroid against its similarity to
\emph{plain-background} and cluttered-background object centroids
(full table in the supplementary material). With full context the
descriptor is far closest to its own category; for the queries that
break, $d_0$ reverses this. The own$-$plain margin flips sign
($+0.25\!\to\!-0.02$), and the descriptor is more similar to plain
than to cluttered backgrounds, so the driver is plainness, not a
generic collapse. The mask-pooled descriptor never pools the gray
fill, but the object's tokens are contextualized by attention over it;
stripped of scene they take on a plain-surround signature. The single
most frequent flip target is \emph{kite}, one of COCO's two
plainest-background categories, absorbing a third of all $d_0$ flips,
and the pull holds across all ten backbones (broken descriptors closer
to plain than cluttered backgrounds in every case; plain overtakes the
own category in seven of ten).

\subsection{Diagnostic Validity}
\label{sec:validity}

\begin{table}[t]
\centering
\small
\setlength{\tabcolsep}{3pt}
\begin{tabular*}{\columnwidth}{@{\extracolsep{\fill}}llcc@{}}
\toprule
Perturbation & What it tests & $\rho$ & mean shift \\
\midrule
Zero-fill        & fill operator / OOD edge   & $0.97$ & n/a \\
Blur-fill        & fill operator / stats & $0.94$ & $5.4$ \\
Shuffle-fill     & fill operator / layout & $0.92$ & $4.5$ \\
Alpha-feather    & sharp-edge tokenization    & $1.00$ & $+1.2$ \\
Boundary jitter ($1^\circ$) & imperfect masks   & $1.00$ & $0.35$ \\
Max-pooling      & pooling operator           & $0.96$ & $2.6$ \\
Attn.\ pooling & pooling operator             & $0.99$ & $2.5$ \\
Retrieval\,$\to$\,classif. & retrieval readout & $0.87$ & n/a \\
\bottomrule
\end{tabular*}
\caption{\textbf{The \dofull{} ranking is robust.} Spearman $\rho$
between the ten-backbone \dofull{} ranking under gray fill and under
each perturbation, with mean absolute \dofull{} shift (pp); the
pooling-operator rows compare max/attentive against the mean-pooling
ranking under an otherwise identical protocol. The
ranking is highly stable across fill choice (gray, zero, blur,
pixel-shuffle), edge softness, realistic mask error, pooling operator
(mean vs.\ max vs.\ attentive), and a swap of the retrieval readout for
a trained classifier (last row). ``n/a'': the perturbation changes the
metric's scale, so only the ranking is comparable.}
\label{tab:validity}
\end{table}

Where \cref{sec:breadth,sec:stratified} measure and \cref{sec:instre}
deploys, this subsection checks the diagnostic itself: that \dofull{}
tracks the representation rather than our measurement choices. All
results use the full ten-backbone panel (\cref{tab:validity}).

\paragraph{Stress tests.}
Neither the fill nor the boundary drives the ranking. Zero fill
preserves it ($\rho{=}0.97$) and alpha-feathering the hard cut leaves
it unchanged ($\rho{=}1.00$, $+1.2$\,pp). A Gaussian-blur fill, which
supplies natural local statistics, keeps it largely intact
($\rho{=}0.94$) at the table's largest shift ($5.4$\,pp): the two
strongest self-supervised backbones (DINOv2-G, DINOv3-L) gain
$9$--$12$\,pp retention under blur. A pixel-shuffle fill (exact
color histogram, layout destroyed) removes exactly that gain
($\rho{=}0.92$; $76.2\to62.8$, $72.2\to55.2$), so the blur-specific
rise is low-frequency context leaking through the fill, not a
gray-patch artifact. A fully \emph{natural} counterfactual needs no
fill at all: cut-pasting the object into a random real scene (INSTRE,
SAM3 masks) drops R@1 on all ten backbones, at least as hard as gray on
eight of ten, with no synthetic fill whatsoever. The
read-out matters as little. Max and attentive pooling (parameter-free,
feature-norm weighted) track the mean-pooled ranking
($\rho{=}0.96/0.99$; the mean values move a few points, but the
ranking is stable). Realistic mask error (per-mask $1^\circ$
rotation, median IoU $0.96$) leaves it identical ($\rho{=}1.00$,
$0.35$\,pp; \cref{sec:instre} closes the loop with fully
\emph{predicted} masks), and replacing retrieval entirely with an
$80$-way linear-probe classifier reproduces it ($\rho{=}0.87$;
ConvNeXt-L $62.2$ vs.\ ViT-L $57.7$). Full mechanism analysis and
per-backbone probe results are in the supplementary material.

\subsection{Reducing Context Dependence}
\label{sec:mitigation}

\begin{table}[t]
\centering
\small
\setlength{\tabcolsep}{5pt}
\begin{tabular*}{\columnwidth}{@{\extracolsep{\fill}}llccc@{}}
\toprule
Backbone & Pretr. & \makecell{\dofull\\frozen} & \makecell{\dofull{} +LoRA\\(95\% CI on $\Delta$)} & \makecell{full R@1\\$\Delta$} \\
\midrule
ViT-L      & Sup & 60.2 & \textbf{64.1} {\scriptsize[+2.0,+5.8]} & $+0.3$ \\
SigLIP2-L  & V-L & 46.0 & \textbf{50.5} {\scriptsize[+3.1,+6.0]} & $+0.5$ \\
EVA-02-MIM & MIM & 27.8 & \textbf{43.6} {\scriptsize[+13.8,+17.8]} & $+6.7$ \\
\bottomrule
\end{tabular*}
\caption{\textbf{Aperture-conditioned LoRA reduces context dependence.}
Object-only retention \dofull{} (\%) on COCO \texttt{val2017} (a
$36$K-object gallery fully disjoint from the $25$K adapter-training
images), before and after a rank-8 adapter, three pretraining
paradigms. Brackets: paired query-bootstrap 95\% CI on the gain
($10^4$ resamples); all three exclude zero. Full-image R@1 is
preserved or improved.}
\label{tab:lora}
\end{table}

A diagnostic is more useful if the property it measures can be moved by a
targeted intervention. We demonstrate, on three backbones spanning the
supervised, vision--language, and masked-image-modeling paradigms, that
a minimal adapter reduces context dependence without harming full-image
accuracy and, under genuine instance retrieval (\cref{sec:instre}),
improves it.

\begin{figure}[t]
  \centering
  \includegraphics[width=\linewidth]{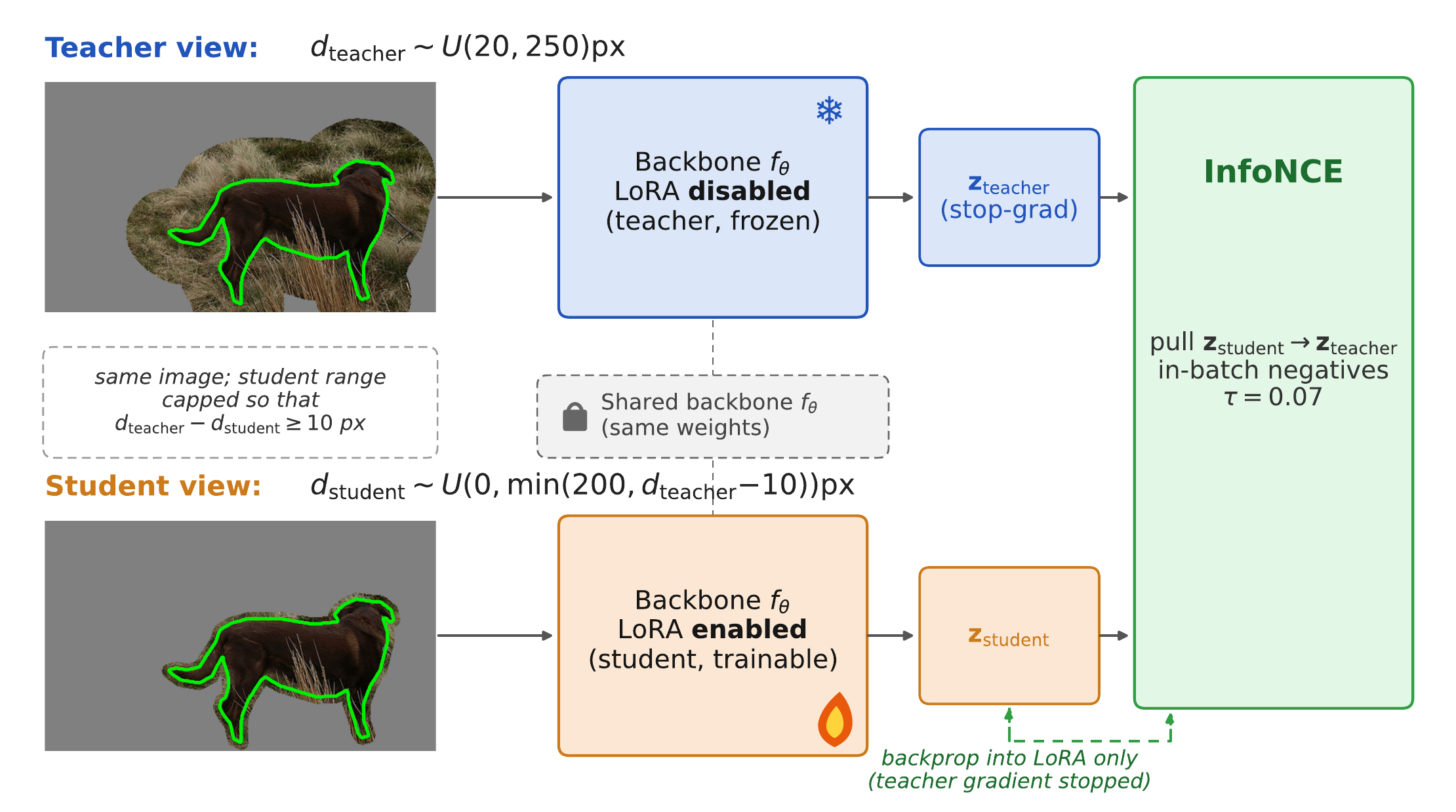}
  \caption{\textbf{Aperture distillation.} A single frozen backbone with
  LoRA toggled off/on produces the teacher (wide aperture) and student
  (tight aperture) views of the same image; an InfoNCE loss pulls the
  student descriptor toward the (gradient-stopped) teacher descriptor,
  updating only the LoRA weights.}
  \label{fig:adapter}
\end{figure}

\paragraph{Recipe.}
The adapter is a rank-8 LoRA~\cite{hu2022lora} on the vision backbone's
attention projections (\texttt{attn.qkv}; separate
\texttt{q/k/v\_proj} on EVA-02-MIM),
trained for one epoch on 25K COCO \texttt{train2017}
images. Adapted models are evaluated only on COCO \texttt{val2017},
INSTRE, and ILIAS, never on the \texttt{train2017} gallery, so no
training image appears in any evaluation gallery. Training is a teacher--student
\emph{aperture distillation} on a single frozen backbone
(\cref{fig:adapter}). The teacher
(LoRA disabled) encodes a wide aperture, $d_{\text{teacher}}\sim
U(20,250)$\,px, and the student (LoRA enabled) a tight one,
$d_{\text{student}}\sim U(0,\min(200,\,d_{\text{teacher}}{-}10))$\,px,
both sampled per image. An InfoNCE loss pulls the
student's global descriptor toward the teacher's (in-batch negatives,
temperature $0.07$, teacher gradient stopped). The outside-mask fill is
randomized over gray, zero, blur, and noise so the adapter does not
overfit one fill (full hyperparameters in the supplementary material).
The objective acts on the \emph{global} descriptor, yet the adapter
improves retrieval under mean, CLS, and attention pooling alike
(\cref{sec:instre}), indicating broad context-robustness rather than a
tuned readout. At $<\!0.4\%$ of backbone parameters the gain reflects
the training signal, not added width.

\paragraph{The adapter reduces context dependence.}
On the three backbones (\cref{tab:lora}), the adapter raises object-only
retention by $+3.9$ to $+15.8$\,pp, every gain with a paired-bootstrap
$95\%$ CI excluding zero, while full-image R@1 is preserved or improved.
The adapter \emph{redistributes} the descriptor's reliance away from
context rather than trading away its ceiling.

\subsection{Transfer to Instance Retrieval}
\label{sec:instre}

\begin{table}[t]
\centering
\small
\setlength{\tabcolsep}{4pt}
\begin{tabular*}{\columnwidth}{@{\extracolsep{\fill}}ll cc cc@{}}
\toprule
 & & \multicolumn{2}{c}{Frozen mAP} & \multicolumn{2}{c}{+LoRA mAP} \\
\cmidrule(lr){3-4}\cmidrule(lr){5-6}
Backbone & pool & full & \makecell{ctx.\\rm.} & full & \makecell{ctx.\\rm.} \\
\midrule
ViT-L      & mean & 46.3 & 43.2 & \textbf{49.9} & \textbf{50.2} \\
ViT-L      & CLS  & 54.8 & 48.1 & \textbf{61.7} & \textbf{60.5} \\
SigLIP2-L  & attn & 81.8 & 74.2 & \textbf{86.2} & \textbf{85.4} \\
EVA-02-MIM & avg  & 35.6 & 19.3 & \textbf{47.5} & \textbf{45.2} \\
\bottomrule
\end{tabular*}
\caption{\textbf{Transfer to INSTRE instance retrieval}
($1{,}208$ queries / $27{,}293$ gallery), mAP (\%). Context removal uses
\emph{SAM-predicted} query masks. The COCO-category-trained adapters of
\cref{sec:mitigation} \textbf{improve full-image mAP in every cell}
($+3.6$ to $+11.9$\,pp) and additionally collapse the
context-removal cost (frozen full$\to$ctx.\ vs.\ LoRA full$\to$ctx.).
All mAP values carry query-bootstrap 95\% CIs of half-width $\le1.8$\,pp;
every full-image gain exceeds its CI.}
\label{tab:instre}
\end{table}

The diagnostic and its fix are defined on category-level retrieval
with oracle masks. We first check the frozen \dofull{} ranking on a
third, instance-level dataset across the full ten-backbone panel
(\cref{tab:ilias}), then test the adapters from \cref{sec:mitigation}
with no further tuning, changing \emph{task}, \emph{dataset}, and
\emph{mask source} at once.

\paragraph{Setup.}
We evaluate on INSTRE~\cite{wang2015instre} ($1{,}208$ queries,
$27{,}293$ gallery, cluttered scenes); query-side context removal uses
masks \emph{predicted} by box-prompted SAM3~\cite{sam3}, the gallery is
natural: the deployment regime the oracle stood in for.

\paragraph{Full-image instance retrieval improves.}
On INSTRE the adapter does more than blunt context removal. It makes a
\emph{better instance descriptor}: full-image mAP rises in every cell
of \cref{tab:instre} by $+3.6$ to $+11.9$\,pp, every gain exceeding
its bootstrap CI, while the context-removal cost collapses
(EVA-02-MIM's $16.3$\,pp full$\to$masked drop shrinks to $2.3$\,pp;
SigLIP2-L's $7.6$ to $0.8$).

\begin{table}[t]
\centering
\small
\setlength{\tabcolsep}{3pt}
\begin{tabular*}{\columnwidth}{@{\extracolsep{\fill}}ll cc cc@{}}
\toprule
 & & \multicolumn{2}{c}{R@1} & \multicolumn{2}{c}{mAP} \\
\cmidrule(lr){3-4}\cmidrule(lr){5-6}
Backbone & Pretr. & full & \dofull & full & \dofull \\
\midrule
DINOv2-G      & SSL & 74.1 & \textbf{98.0} & 64.8 & \textbf{94.9} \\
DINOv3-L      & SSL & 79.7 & \textbf{89.6} & 72.4 & \textbf{85.6} \\
ViT-L         & Sup & 71.5 & \textbf{86.3} & 61.5 & \textbf{83.2} \\
ConvNeXt-L    & Sup & 69.0 & \textbf{84.7} & 59.6 & \textbf{78.2} \\
SigLIP2-512   & V-L & 82.0 & \textbf{80.9} & 69.5 & \textbf{73.8} \\
MAE ViT-L$^\dagger$ & MIM & 13.5 & \textbf{78.7} & 9.6 & \textbf{80.9} \\
SigLIP2-L     & V-L & 79.1 & \textbf{78.6} & 65.8 & \textbf{69.9} \\
EVA-02-CLIP & V-L & 77.6 & \textbf{76.0} & 67.1 & \textbf{70.8} \\
EVA-02-MIM  & MIM & 53.7 & \textbf{36.3} & 40.5 & \textbf{33.1} \\
DINO ViT-S/8  & SSL & 51.2 & \textbf{19.8} & 35.7 & \textbf{13.7} \\
\bottomrule
\end{tabular*}
\caption{\textbf{ILIAS instance-level replication} (all ten frozen
backbones, gray fill, same mask-pooled protocol as \cref{tab:breadth}).
Each query is a cluttered positive with a SAM3~\cite{sam3} box-prompted mask,
context-removed and pooled over the object mask exactly as in the
category-level probe; retrieval is against the $4{,}653$ natural
positives with the query image excluded from the gallery. This is
ILIAS-core scale: we do not add the benchmark's $100$M YFCC
distractors, so these numbers measure the diagnostic's transfer, not
ILIAS leaderboard performance. R@1 and mAP (\%) at
full image, and object-only retention (\%) formed as the $d_0$/full
ratio \emph{within each metric} (\dofull{} of \cref{tab:breadth} is the
R@1 column). The backbone
ranking matches COCO (Spearman $\rho{=}0.94$, equal to the
COCO$\leftrightarrow$ADE20K transfer), so the
diagnostic generalizes to a third dataset and to instance-level
retrieval. $^\dagger$MAE's ratio is a floor artifact (near-chance
full-image accuracy), as elsewhere.}
\label{tab:ilias}
\end{table}

\paragraph{ILIAS positive-as-query.}
INSTRE tests one mask source and one dataset; we further check the
category-level probe against a genuinely different instance-retrieval
benchmark, ILIAS~\cite{kordopatiszilos2025ilias}. ILIAS's own queries
are already context-removed by construction. Auditing all $1{,}232$, we
find $17.1\%$ are hard-segmented cutouts (an exactly constant fill
wrapping the object silhouette) against $0.6\%$ of its natural positives
(test in the supplementary material), the rest clean or bare-surface
backgrounds. They therefore
cannot serve as a masking counterfactual; instead
we treat a randomly chosen \emph{cluttered} positive as the query,
mirroring the COCO protocol exactly (\cref{tab:ilias}). Context removal
hurts every backbone, most severely for small objects (retention
about $11\%$ on the weakest), and the adapters from \cref{sec:mitigation}
transfer here too without further tuning: EVA-02-MIM gains $+19.3$
full-image R@1 while its \dofull{} nearly doubles
($36.3\to70.9$); per-backbone deltas with paired CIs appear in
\cref{tab:control}. ILIAS thus confirms both halves of the
diagnostic, the ranking and the adapter's transfer, on a benchmark
built for a different task than the one KAP was designed against.

\paragraph{A matched control isolates the aperture signal.}
The adapter's gain could in principle come from contrastive fine-tuning
on COCO images generically, rather than from the aperture curriculum
specifically. We rule this out with a strictly matched control
(``SimCLR aug.''): identical teacher--student topology (LoRA-disabled
teacher with stopped gradient, LoRA student), identical training
images, InfoNCE objective, and hyperparameters. The \emph{only}
change is the views, which are two standard SimCLR
augmentations~\cite{chen2020simclr} of the full frame instead of a
wide/tight aperture pair.

\begin{table}[t]
\centering
\small
\setlength{\tabcolsep}{5pt}
\textbf{ILIAS R@1 (\%)}\\[2pt]
\begin{tabular*}{\columnwidth}{@{\extracolsep{\fill}}lccc@{}}
\toprule
Backbone & frozen & $\Delta$\,SimCLR aug. & $\Delta$\,KAP\\
\midrule
EVA-02-MIM & 53.7 & $+6.1_{\pm2.6}$ & $\boldsymbol{+19.3}_{\pm2.8}$\\
ViT-L      & 71.5 & $-2.1_{\pm1.4}$ & $+0.1_{\pm1.5}$\\
SigLIP2-L  & 79.1 & $+1.8_{\pm1.9}$ & $+0.9_{\pm1.5}$\\
\bottomrule
\end{tabular*}\\[6pt]
\textbf{INSTRE R@1 (\%)}\\[2pt]
\begin{tabular*}{\columnwidth}{@{\extracolsep{\fill}}lccc@{}}
\toprule
Backbone & frozen & $\Delta$\,SimCLR aug. & $\Delta$\,KAP\\
\midrule
EVA-02-MIM & 81.0 & $+3.6_{\pm1.8}$ & $\boldsymbol{+8.6}_{\pm1.9}$\\
ViT-L      & 86.8 & $-1.1_{\pm1.1}$ & $\boldsymbol{+2.9}_{\pm1.3}$\\
SigLIP2-L  & 98.5 & $-2.3_{\pm1.0}$ & $\boldsymbol{+0.6}_{\pm0.5}$\\
\bottomrule
\end{tabular*}
\caption{\textbf{The aperture curriculum, not generic contrastive
fine-tuning, drives the gain.} Full-image R@1 on natural queries
(frozen baseline), with the \emph{change} produced by the matched
augmentation-views control ($\Delta$\,SimCLR aug.) and by the aperture
adapter ($\Delta$\,KAP); subscripts are paired query-bootstrap $95\%$
CI half-widths on each $\Delta$; bold marks KAP improvements whose CI
excludes zero. Head-to-head,
KAP exceeds the control with paired CIs excluding zero in five of the
six cells (SigLIP2-L on ILIAS is a tie). ILIAS uses the mask-pooled
positive-as-query protocol of \cref{tab:ilias}; INSTRE uses
\cref{tab:instre}'s full-image protocol with each backbone's own
pooling, reported here as R@1.}
\label{tab:control}
\end{table}

The control changes little or hurts: it significantly degrades frozen
ViT-L on ILIAS and frozen SigLIP2-L on INSTRE, and its one clear gain
(EVA-02-MIM) is less than a third of KAP's on the same backbone. Under
\cref{tab:instre}'s own mAP metric the contrast is starker still: the
control leaves ViT-L below frozen ($46.3\to44.9$ vs.\ KAP's $49.9$)
and collapses SigLIP2-L ($81.8\to62.6$ vs.\ $86.2$), so the
\cref{tab:instre} gains themselves are attributable to the aperture
curriculum. A conventional \emph{symmetric}
SimCLR variant (both branches trainable, no teacher) shows the same
qualitative pattern (supplementary material).

\paragraph{Generalization axes.}
The transfer crosses four axes at once: task granularity
(category $\to$ instance), dataset (COCO $\to$ INSTRE and ILIAS), mask
source (oracle $\to$ box-prompted SAM3), and
descriptor (mean / CLS / attention pooling), ruling out the concern
that the fix is ``trained to its own probe.''

\section{Conclusion}
\label{sec:conclusion}

We introduced the Keyhole Aperture Probe, a counterfactual diagnostic
that reduces the cost of context removal to a single scalar by fixing
the pooling region and intervening only outside it. The pooled object
region has already integrated its surround: replacing the context
hurts whether the replacement is a neutral fill or a different real
scene, and for objects that break, the descriptor crosses over to
plain-background images rather than its own category. Object-only is
not context-free. Across ten backbones this cost varies widely, is
strongly graded by object size, and \emph{does not require ViT
self-attention}. The score is also \emph{actionable}: a lightweight
aperture-conditioned adapter improves full-image instance retrieval on
INSTRE, transferring to ILIAS with accuracy improved or
preserved, while closing the context-removal gap, and a matched
control confirms the gain is specific to the aperture curriculum. We
release the protocol, code, and adapter checkpoints at
\url{https://github.com/bmeivar/kap} so practitioners can measure, and if needed reduce, this trade-off
for new backbones.

\paragraph{Limitations.}
KAP measures how much a descriptor \emph{depends} on context, not
whether that context is spurious: at category level part of the
\dofull{} gap is lost predictive scene evidence, which is why we also
report instance-level results. \dofull{} is a ratio that flatters
low-accuracy backbones (MAE) and should be read with full-image R@1 or
AUC. We claim an architecture effect, not an objective effect: the
panel confounds objective with scale and data, and ConvNeXt is our only
CNN. The adapter is a Transformer-only proof of concept, and the
diagnostic uses oracle masks.

\paragraph{Acknowledgement.}
Part of this research was supported by ISF grant 2132/23.

{
    \small
    \bibliographystyle{ieeenat_fullname}
    \bibliography{main}

\begin{thebibliography}{37}
\providecommand{\natexlab}[1]{#1}
\providecommand{\url}[1]{\texttt{#1}}
\expandafter\ifx\csname urlstyle\endcsname\relax
  \providecommand{\doi}[1]{doi: #1}\else
  \providecommand{\doi}{doi: \begingroup \urlstyle{rm}\Url}\fi

\bibitem[Aravanis et~al.(2026)Aravanis, Stojni{\'c}, Psomas, Komodakis, and
  Tolias]{aravanis2026rns}
Tilemachos Aravanis, Vladan Stojni{\'c}, Bill Psomas, Nikos Komodakis, and
  Giorgos Tolias.
\newblock Retrieve and segment: Are a few examples enough to bridge the
  supervision gap in open-vocabulary segmentation?
\newblock In \emph{CVPR}, pages 27621--27632, 2026.

\bibitem[Cao et~al.(2020)Cao, Ara{\'u}jo, and Sim]{cao2020delg}
Bingyi Cao, Andr{\'e} Ara{\'u}jo, and Jack Sim.
\newblock Unifying deep local and global features for image search.
\newblock In \emph{Computer Vision -- ECCV 2020}, pages 726--743. Springer,
  2020.

\bibitem[Carion et~al.(2026)Carion, Gustafson, Hu, Debnath, Hu, Suris, Ryali,
  Alwala, Khedr, Huang, Lei, Ma, Guo, Kalla, Marks, Greer, Wang, Sun,
  R{\"a}dle, Afouras, Mavroudi, Xu, Wu, Zhou, Momeni, Hazra, Ding, Vaze,
  Porcher, Li, Li, Kamath, Cheng, Doll{\'a}r, Ravi, Saenko, Zhang, and
  Feichtenhofer]{sam3}
Nicolas Carion, Laura Gustafson, Yuan-Ting Hu, Shoubhik Debnath, Ronghang Hu,
  Didac Suris, Chaitanya Ryali, Kalyan~Vasudev Alwala, Haitham Khedr, Andrew
  Huang, Jie Lei, Tengyu Ma, Baishan Guo, Arpit Kalla, Markus Marks, Joseph
  Greer, Meng Wang, Peize Sun, Roman R{\"a}dle, Triantafyllos Afouras,
  Effrosyni Mavroudi, Katherine Xu, Tsung-Han Wu, Yu Zhou, Liliane Momeni,
  Rishi Hazra, Shuangrui Ding, Sagar Vaze, Francois Porcher, Feng Li, Siyuan
  Li, Aishwarya Kamath, Ho~Kei Cheng, Piotr Doll{\'a}r, Nikhila Ravi, Kate
  Saenko, Pengchuan Zhang, and Christoph Feichtenhofer.
\newblock {SAM} 3: Segment anything with concepts.
\newblock In \emph{ICLR}, 2026.

\bibitem[Caron et~al.(2021)Caron, Touvron, Misra, J{\'e}gou, Mairal,
  Bojanowski, and Joulin]{caron2021dino}
Mathilde Caron, Hugo Touvron, Ishan Misra, Herv{\'e} J{\'e}gou, Julien Mairal,
  Piotr Bojanowski, and Armand Joulin.
\newblock Emerging properties in self-supervised vision transformers.
\newblock In \emph{ICCV}, 2021.

\bibitem[Chen et~al.(2020)Chen, Kornblith, Norouzi, and Hinton]{chen2020simclr}
Ting Chen, Simon Kornblith, Mohammad Norouzi, and Geoffrey Hinton.
\newblock A simple framework for contrastive learning of visual
  representations.
\newblock In \emph{ICML}, 2020.

\bibitem[Didolkar et~al.(2025)Didolkar, Zadaianchuk, Goyal, Mozer, Bengio,
  Martius, and Seitzer]{didolkar2024ftdinosaur}
Aniket Didolkar, Andrii Zadaianchuk, Anirudh Goyal, Michael Mozer, Yoshua
  Bengio, Georg Martius, and Maximilian Seitzer.
\newblock On the transfer of object-centric representation learning.
\newblock In \emph{ICLR}, 2025.

\bibitem[Ding et~al.(2022)Ding, Xue, Xia, and Dai]{ding2022zegformer}
Jian Ding, Nan Xue, Gui-Song Xia, and Dengxin Dai.
\newblock Decoupling zero-shot semantic segmentation.
\newblock In \emph{CVPR}, 2022.

\bibitem[Fang et~al.(2024)Fang, Sun, Wang, Huang, Wang, and Cao]{fang2024eva02}
Yuxin Fang, Quan Sun, Xinggang Wang, Tiejun Huang, Xinlong Wang, and Yue Cao.
\newblock {EVA-02}: A visual representation for neon genesis.
\newblock \emph{Image and Vision Computing}, 149:\penalty0 105171, 2024.

\bibitem[Fong and Vedaldi(2017)]{fong2017meaningful}
Ruth~C. Fong and Andrea Vedaldi.
\newblock Interpretable explanations of black boxes by meaningful perturbation.
\newblock In \emph{Proceedings of the IEEE International Conference on Computer
  Vision (ICCV)}, 2017.

\bibitem[Geirhos et~al.(2020)Geirhos, Jacobsen, Michaelis, Zemel, Brendel,
  Bethge, and Wichmann]{geirhos2020shortcut}
Robert Geirhos, J{\"o}rn-Henrik Jacobsen, Claudio Michaelis, Richard Zemel,
  Wieland Brendel, Matthias Bethge, and Felix~A. Wichmann.
\newblock Shortcut learning in deep neural networks.
\newblock \emph{Nature Machine Intelligence}, 2\penalty0 (11):\penalty0
  665--673, 2020.

\bibitem[He et~al.(2022)He, Chen, Xie, Li, Doll{\'a}r, and Girshick]{he2022mae}
Kaiming He, Xinlei Chen, Saining Xie, Yanghao Li, Piotr Doll{\'a}r, and Ross
  Girshick.
\newblock Masked autoencoders are scalable vision learners.
\newblock In \emph{CVPR}, 2022.

\bibitem[Hu et~al.(2022)Hu, Shen, Wallis, Allen-Zhu, Li, Wang, Wang, and
  Chen]{hu2022lora}
Edward~J Hu, Yelong Shen, Phillip Wallis, Zeyuan Allen-Zhu, Yuanzhi Li, Shean
  Wang, Lu Wang, and Weizhu Chen.
\newblock {LoRA}: Low-rank adaptation of large language models.
\newblock In \emph{ICLR}, 2022.

\bibitem[Jia et~al.(2023)Jia, Liu, and Huang]{jia2023boqsa}
Baoxiong Jia, Yu Liu, and Siyuan Huang.
\newblock Improving object-centric learning with query optimization.
\newblock In \emph{ICLR}, 2023.

\bibitem[Kirillov et~al.(2019)Kirillov, He, Girshick, Rother, and
  Doll{\'a}r]{kirillov2019panoptic}
Alexander Kirillov, Kaiming He, Ross Girshick, Carsten Rother, and Piotr
  Doll{\'a}r.
\newblock Panoptic segmentation.
\newblock In \emph{Proceedings of the IEEE/CVF Conference on Computer Vision
  and Pattern Recognition (CVPR)}, pages 9404--9413, 2019.

\bibitem[Kordopatis-Zilos et~al.(2025)Kordopatis-Zilos, Stojni\'{c}, Manko,
  \v{S}uma, Ypsilantis, Efthymiadis, Laskar, Matas, Chum, and
  Tolias]{kordopatiszilos2025ilias}
Giorgos Kordopatis-Zilos, Vladan Stojni\'{c}, Anna Manko, Pavel \v{S}uma,
  Nikolaos-Antonios Ypsilantis, Nikos Efthymiadis, Zakaria Laskar, Ji\v{r}\'{i}
  Matas, Ond\v{r}ej Chum, and Giorgos Tolias.
\newblock {ILIAS}: Instance-level image retrieval at scale.
\newblock In \emph{CVPR}, 2025.

\bibitem[Korfhage et~al.(2025)Korfhage, M{\"u}hling, and
  Freisleben]{korfhage2024searchanything}
Nikolaus Korfhage, Markus M{\"u}hling, and Bernd Freisleben.
\newblock Search anything: Segmentation-based similarity search via region
  prompts.
\newblock \emph{Multimedia Tools and Applications}, 84\penalty0 (27):\penalty0
  32593--32618, 2025.

\bibitem[Li et~al.(2025)Li, Cheng, Feng, Liu, and Wang]{li2025maskadapter}
Yongkang Li, Tianheng Cheng, Bin Feng, Wenyu Liu, and Xinggang Wang.
\newblock Mask-adapter: The devil is in the masks for open-vocabulary
  segmentation.
\newblock In \emph{CVPR}, 2025.

\bibitem[Liang et~al.(2023)Liang, Wu, Dai, Li, Zhao, Zhang, Zhang, Vajda, and
  Marculescu]{liang2023ovseg}
Feng Liang, Bichen Wu, Xiaoliang Dai, Kunpeng Li, Yinan Zhao, Hang Zhang,
  Peizhao Zhang, Peter Vajda, and Diana Marculescu.
\newblock Open-vocabulary semantic segmentation with mask-adapted {CLIP}.
\newblock In \emph{CVPR}, 2023.

\bibitem[Lin et~al.(2014)Lin, Maire, Belongie, Hays, Perona, Ramanan,
  Doll{\'a}r, and Zitnick]{lin2014coco}
Tsung-Yi Lin, Michael Maire, Serge Belongie, James Hays, Pietro Perona, Deva
  Ramanan, Piotr Doll{\'a}r, and C.~Lawrence Zitnick.
\newblock Microsoft {COCO}: Common objects in context.
\newblock In \emph{Computer Vision -- ECCV 2014}, pages 740--755. Springer,
  2014.

\bibitem[Liu et~al.(2022)Liu, Mao, Wu, Feichtenhofer, Darrell, and
  Xie]{liu2022convnext}
Zhuang Liu, Hanzi Mao, Chao-Yuan Wu, Christoph Feichtenhofer, Trevor Darrell,
  and Saining Xie.
\newblock A {ConvNet} for the 2020s.
\newblock In \emph{CVPR}, 2022.

\bibitem[Locatello et~al.(2020)Locatello, Weissenborn, Unterthiner, Mahendran,
  Heigold, Uszkoreit, Dosovitskiy, and Kipf]{locatello2020slotattention}
Francesco Locatello, Dirk Weissenborn, Thomas Unterthiner, Aravindh Mahendran,
  Georg Heigold, Jakob Uszkoreit, Alexey Dosovitskiy, and Thomas Kipf.
\newblock Object-centric learning with slot attention.
\newblock In \emph{NeurIPS}, 2020.

\bibitem[Noh et~al.(2017)Noh, Araujo, Sim, Weyand, and Han]{noh2017delf}
Hyeonwoo Noh, Andre Araujo, Jack Sim, Tobias Weyand, and Bohyung Han.
\newblock Large-scale image retrieval with attentive deep local features.
\newblock In \emph{Proceedings of the IEEE International Conference on Computer
  Vision (ICCV)}, pages 3456--3465, 2017.

\bibitem[Oliva and Torralba(2007)]{oliva2007role}
Aude Oliva and Antonio Torralba.
\newblock The role of context in object recognition.
\newblock \emph{Trends in Cognitive Sciences}, 11\penalty0 (12):\penalty0
  520--527, 2007.

\bibitem[Oquab et~al.(2024)Oquab, Darcet, Moutakanni, Vo, Szafraniec, Khalidov,
  Fernandez, Haziza, Massa, El-Nouby, Assran, Ballas, Galuba, Howes, Huang, Li,
  Misra, Rabbat, Sharma, Synnaeve, Xu, J{\'e}gou, Mairal, Labatut, Joulin, and
  Bojanowski]{oquab2023dinov2}
Maxime Oquab, Timoth{\'e}e Darcet, Th{\'e}o Moutakanni, Huy~V. Vo, Marc
  Szafraniec, Vasil Khalidov, Pierre Fernandez, Daniel Haziza, Francisco Massa,
  Alaaeldin El-Nouby, Mahmoud Assran, Nicolas Ballas, Wojciech Galuba, Russell
  Howes, Po-Yao Huang, Shang-Wen Li, Ishan Misra, Michael Rabbat, Vasu Sharma,
  Gabriel Synnaeve, Hu Xu, Herv{\'e} J{\'e}gou, Julien Mairal, Patrick Labatut,
  Armand Joulin, and Piotr Bojanowski.
\newblock {DINOv2}: Learning robust visual features without supervision.
\newblock \emph{Transactions on Machine Learning Research}, 2024.

\bibitem[Radenovi{\'c} et~al.(2019)Radenovi{\'c}, Tolias, and
  Chum]{radenovic2018gem}
Filip Radenovi{\'c}, Giorgos Tolias, and Ond{\v{r}}ej Chum.
\newblock Fine-tuning {CNN} image retrieval with no human annotation.
\newblock \emph{IEEE Transactions on Pattern Analysis and Machine
  Intelligence}, 41\penalty0 (7):\penalty0 1655--1668, 2019.

\bibitem[Seitzer et~al.(2023)Seitzer, Horn, Zadaianchuk, Zietlow, Xiao,
  Simon-Gabriel, He, Zhang, Sch{\"o}lkopf, Brox, and
  Locatello]{seitzer2023dinosaur}
Maximilian Seitzer, Max Horn, Andrii Zadaianchuk, Dominik Zietlow, Tianjun
  Xiao, Carl-Johann Simon-Gabriel, Tong He, Zheng Zhang, Bernhard
  Sch{\"o}lkopf, Thomas Brox, and Francesco Locatello.
\newblock Bridging the gap to real-world object-centric learning.
\newblock In \emph{ICLR}, 2023.

\bibitem[Shlapentokh-Rothman et~al.(2024)Shlapentokh-Rothman, Blume, Xiao, Wu,
  TV, Tao, Lee, Torres, Wang, and Hoiem]{shlapentokh2024region}
Michal Shlapentokh-Rothman, Ansel Blume, Yao Xiao, Yuqun Wu, Sethuraman TV,
  Heyi Tao, Jae~Yong Lee, Wilfredo Torres, Yu-Xiong Wang, and Derek Hoiem.
\newblock Region-based representations revisited.
\newblock In \emph{CVPR}, 2024.

\bibitem[Sim{\'e}oni et~al.(2025)Sim{\'e}oni, Vo, Seitzer, Baldassarre, Oquab,
  Jose, Khalidov, Szafraniec, Yi, Ramamonjisoa, Massa, Haziza, Wehrstedt, Wang,
  Darcet, Moutakanni, Sentana, Roberts, Vedaldi, Tolan, Brandt, Couprie,
  Mairal, J{\'e}gou, Labatut, and Bojanowski]{simeoni2025dinov3}
Oriane Sim{\'e}oni, Huy~V. Vo, Maximilian Seitzer, Federico Baldassarre, Maxime
  Oquab, Cijo Jose, Vasil Khalidov, Marc Szafraniec, Seungeun Yi, Micha{\"e}l
  Ramamonjisoa, Francisco Massa, Daniel Haziza, Luca Wehrstedt, Jianyuan Wang,
  Timoth{\'e}e Darcet, Th{\'e}o Moutakanni, Leonel Sentana, Claire Roberts,
  Andrea Vedaldi, Jamie Tolan, John Brandt, Camille Couprie, Julien Mairal,
  Herv{\'e} J{\'e}gou, Patrick Labatut, and Piotr Bojanowski.
\newblock {DINOv3}, 2025.

\bibitem[Tolias et~al.(2016)Tolias, Sicre, and J{\'e}gou]{tolias2016rmac}
Giorgos Tolias, Ronan Sicre, and Herv{\'e} J{\'e}gou.
\newblock Particular object retrieval with integral max-pooling of {CNN}
  activations.
\newblock In \emph{International Conference on Learning Representations
  (ICLR)}, 2016.

\bibitem[Torralba(2003)]{torralba2003context}
Antonio Torralba.
\newblock Contextual priming for object detection.
\newblock \emph{International Journal of Computer Vision}, 53\penalty0
  (2):\penalty0 169--191, 2003.

\bibitem[Tschannen et~al.(2025)Tschannen, Gritsenko, Wang, Naeem,
  Alabdulmohsin, Parthasarathy, Evans, Beyer, Xia, Mustafa, H{\'e}naff,
  Harmsen, Steiner, and Zhai]{tschannen2025siglip2}
Michael Tschannen, Alexey Gritsenko, Xiao Wang, Muhammad~Ferjad Naeem, Ibrahim
  Alabdulmohsin, Nikhil Parthasarathy, Talfan Evans, Lucas Beyer, Ye Xia, Basil
  Mustafa, Olivier H{\'e}naff, Jeremiah Harmsen, Andreas Steiner, and Xiaohua
  Zhai.
\newblock {SigLIP} 2: Multilingual vision-language encoders with improved
  semantic understanding, localization, and dense features.
\newblock \emph{arXiv preprint arXiv:2502.14786}, 2025.

\bibitem[Wang and Jiang(2015)]{wang2015instre}
Shuang Wang and Shuqiang Jiang.
\newblock {INSTRE}: A new benchmark for instance-level object retrieval and
  recognition.
\newblock \emph{ACM Transactions on Multimedia Computing, Communications, and
  Applications}, 11\penalty0 (3):\penalty0 1--21, 2015.

\bibitem[Xiao et~al.(2021)Xiao, Engstrom, Ilyas, and Madry]{xiao2021noise}
Kai Xiao, Logan Engstrom, Andrew Ilyas, and Aleksander Madry.
\newblock Noise or signal: The role of image backgrounds in object recognition.
\newblock In \emph{ICLR}, 2021.

\bibitem[Xu et~al.(2022)Xu, Zhang, Wei, Lin, Cao, Hu, and
  Bai]{xu2022simplebaseline}
Mengde Xu, Zheng Zhang, Fangyun Wei, Yutong Lin, Yue Cao, Han Hu, and Xiang
  Bai.
\newblock A simple baseline for open-vocabulary semantic segmentation with
  pre-trained vision-language model.
\newblock In \emph{ECCV}, 2022.

\bibitem[Yu et~al.(2023)Yu, He, Deng, Shen, and Chen]{yu2023fcclip}
Qihang Yu, Ju He, Xueqing Deng, Xiaohui Shen, and Liang-Chieh Chen.
\newblock Convolutions die hard: Open-vocabulary segmentation with single
  frozen convolutional {CLIP}.
\newblock In \emph{NeurIPS}, 2023.

\bibitem[Zhai et~al.(2023)Zhai, Mustafa, Kolesnikov, and Beyer]{zhai2023siglip}
Xiaohua Zhai, Basil Mustafa, Alexander Kolesnikov, and Lucas Beyer.
\newblock Sigmoid loss for language image pre-training.
\newblock In \emph{Proceedings of the IEEE/CVF International Conference on
  Computer Vision (ICCV)}, pages 11975--11986, 2023.

\bibitem[Zhou et~al.(2017)Zhou, Zhao, Puig, Fidler, Barriuso, and
  Torralba]{zhou2017ade20k}
Bolei Zhou, Hang Zhao, Xavier Puig, Sanja Fidler, Adela Barriuso, and Antonio
  Torralba.
\newblock Scene parsing through {ADE20K} dataset.
\newblock In \emph{Proceedings of the IEEE Conference on Computer Vision and
  Pattern Recognition (CVPR)}, pages 5122--5130, 2017.

\end{thebibliography}
}

\clearpage
\setcounter{section}{0}\renewcommand{\thesection}{\Alph{section}}
\setcounter{table}{0}\renewcommand{\thetable}{S\arabic{table}}
\setcounter{figure}{0}\renewcommand{\thefigure}{S\arabic{figure}}
\twocolumn[\centering{\Large\bfseries Supplementary Material}\vspace{14pt}]

\section{Backbone specifications and feature extraction}
\label{sup:backbones}
\cref{tab:backbones} gives the exact checkpoint, input resolution, and
feature grid for every backbone in the panel, so the protocol can be
reproduced without ambiguity. All checkpoints are \texttt{timm}
identifiers loaded with pretrained weights; no backbone is fine-tuned
except where an adapter is explicitly applied (\cref{sup:lora}).

\begin{table}[t]
\centering\footnotesize
\setlength{\tabcolsep}{2pt}
\begin{tabular*}{\columnwidth}{@{\extracolsep{\fill}}llcc@{}}
\toprule
Backbone & \texttt{timm} checkpoint & res. & grid \\
\midrule
DINOv2-G/14  & \texttt{vit\_giant\_patch14\_dinov2}          & 518 & $37^2$ \\
             & \texttt{\,\,.lvd142m}                          &     &        \\
ConvNeXt-L   & \texttt{convnext\_large}                       & 384 & $12^2$ \\
             & \texttt{\,\,.fb\_in22k\_ft\_in1k\_384}         &     &        \\
DINOv3-L     & \texttt{vit\_large\_patch16\_dinov3}           & 256 & $16^2$ \\
ViT-L        & \texttt{vit\_large\_patch16\_384}              & 384 & $24^2$ \\
             & \texttt{\,\,.augreg\_in21k\_ft\_in1k}          &     &        \\
MAE ViT-L    & \texttt{vit\_large\_patch16\_224.mae}          & 224 & $14^2$ \\
SigLIP2-512  & \texttt{vit\_so400m\_patch16\_siglip\_512}     & 512 & $32^2$ \\
             & \texttt{\,\,.v2\_webli}                        &     &        \\
SigLIP2-L    & \texttt{vit\_large\_patch16\_siglip\_384}      & 384 & $24^2$ \\
             & \texttt{\,\,.v2\_webli}                        &     &        \\
EVA-02-CLIP  & \texttt{eva02\_large\_patch14\_clip\_224}      & 224 & $16^2$ \\
             & \texttt{\,\,.merged2b}                         &     &        \\
DINO ViT-S/8 & \texttt{vit\_small\_patch8\_224.dino}          & 224 & $28^2$ \\
EVA-02-MIM   & \texttt{eva02\_large\_patch14\_224}            & 224 & $16^2$ \\
             & \texttt{\,\,.mim\_in22k}                       &     &        \\
\bottomrule
\end{tabular*}
\caption{\textbf{Panel specifications.} Exact \texttt{timm} checkpoint,
square input resolution (px), and resulting dense feature grid for each
of the ten backbones. Grid is $(\text{res}/\text{patch})^2$ for ViTs and
the stride-$32$ stage-4 map for ConvNeXt-L. The SigLIP2 pair differs
only in resolution ($512$ vs.\ $384$, i.e.\ a $32^2$ vs.\ $24^2$ grid),
so the $5$\,pp \dofull{} gap between them in the main paper's panel is
a resolution effect: a finer grid leaves a small object more tokens to
pool.}
\label{tab:backbones}
\end{table}

\paragraph{Order of operations.} The aperture fill is applied to the
\emph{original-resolution} image, before the backbone's preprocessing
transform, so the fill is resized by exactly the same interpolation as
the image content and no fill-versus-content resampling mismatch is
introduced. For pooling we \emph{upsample the dense feature map
bilinearly to mask resolution} rather than downsampling the mask to the
feature grid; this matters for small objects, whose masks can vanish
entirely under nearest-neighbour downsampling to a $14^2$ or $16^2$
grid. Precision is fp16 for all backbones except DINOv3-L (bf16).

\section{ADE20K cross-dataset replication}
\label{sup:ade20k}
The full ten-backbone panel re-run on ADE20K (scene-parsing instances)
under the identical protocol, referenced from the main paper's
``Cross-dataset replication'' paragraph. The diagnostic transfers: the
backbone ranking is preserved (Spearman $\rho{=}0.94$, $p<10^{-3}$
against COCO \dofull{}), the architecture dissociation reproduces---a
convolutional ConvNeXt-L is again more context-resistant than ViT-L
($68$ vs.\ $60$, $+8$\,pp)---and the extremes agree, with DINOv2-G and
ConvNeXt-L highest and DINO ViT-S/8 and EVA-02-MIM lowest on both
datasets (\cref{tab:ade20k}).

\begin{table}[t]
\centering\small
\setlength{\tabcolsep}{5pt}
\begin{tabular*}{\columnwidth}{@{\extracolsep{\fill}}lccc c@{}}
\toprule
 & \multicolumn{3}{c}{ADE20K} & COCO \\
\cmidrule(lr){2-4}\cmidrule(lr){5-5}
Backbone & $d_0$ & full & \dofull & \dofull \\
\midrule
DINOv2-G       & 45.0 & 65.0 & \textbf{69} & 67 \\
ConvNeXt-L     & 31.6 & 46.7 & \textbf{68} & 66 \\
DINOv3-L       & 39.1 & 63.1 & \textbf{62} & 60 \\
ViT-L          & 34.4 & 57.3 & \textbf{60} & 59 \\
MAE ViT-L      & 18.8 & 30.5 & \textbf{62}\rlap{$^\dagger$} & 57 \\
SigLIP2-512    & 28.3 & 57.1 & \textbf{50} & 55 \\
SigLIP2-L      & 32.1 & 64.1 & \textbf{50} & 50 \\
EVA-02-CLIP    & 29.9 & 56.4 & \textbf{53} & 48 \\
DINO ViT-S/8   & 12.4 & 50.1 & \textbf{25} & 32 \\
EVA-02-MIM     & 12.3 & 50.3 & \textbf{25} & 28 \\
\bottomrule
\end{tabular*}
\caption{\textbf{ADE20K cross-dataset replication} (all ten backbones,
gray fill), R@1 (\%) at $d_0$ and full image with object-only retention
\dofull{} (\%), beside COCO \dofull{} for reference. The ranking is
preserved across datasets (Spearman $\rho{=}0.94$), and ConvNeXt-L again
exceeds ViT-L ($+8$\,pp). $^\dagger$MAE's ratio is unreliable (floor
full-image accuracy), as on COCO.}
\label{tab:ade20k}
\end{table}

\section{Recovery curves}
\label{sup:recovery}
\begin{figure*}[t]
  \centering
  \includegraphics[width=0.80\linewidth]{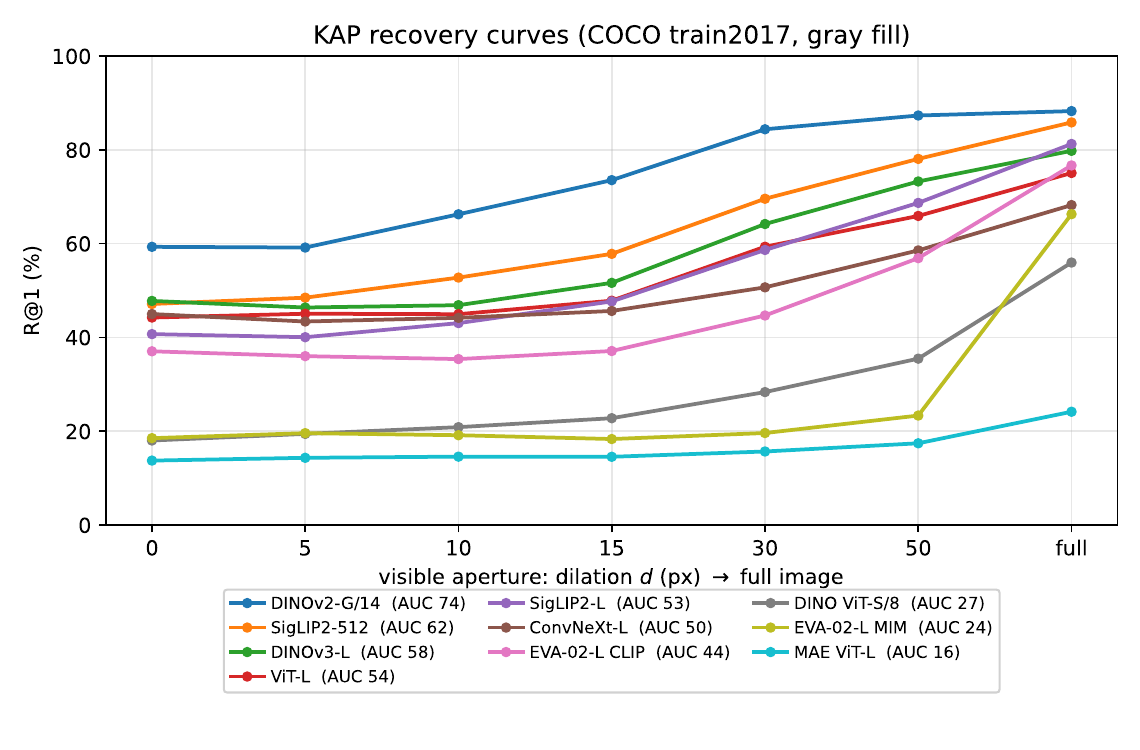}
  \caption{\textbf{KAP recovery curves.} R@1 vs.\ visible aperture
  ($d{=}0,5,\dots,50$\,px\,$\to$\,full image, gray fill) for all ten
  backbones on COCO \texttt{train2017}, with the area under each curve
  (AUC = mean R@1 over the sweep, \%) in the legend, sorted by AUC.
  All ten recover gradually---none regains full accuracy until most
  context is restored---and the spread shows how differently the
  backbones rely on context.}
  \label{fig:recovery}
\end{figure*}
\Cref{fig:recovery} plots R@1 against the visible aperture for all ten
backbones (referenced from the main paper's ``Recovery curves''
paragraph). Recovery is gradual and the curves separate cleanly; the
legend's AUC values match Tab.~1 of the main paper.

\section{Stratified effects}
\label{sup:stratified}
\begin{figure*}[t]
  \centering
  \includegraphics[width=0.72\linewidth]{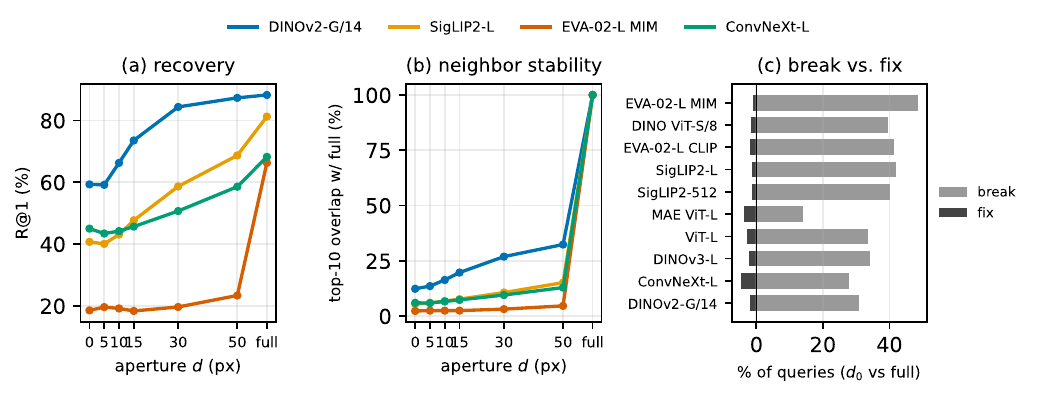}
  \caption{\textbf{Stratified effects} (one backbone per paradigm).
  (a) Retrieval recovery vs.\ aperture; (b) top-$10$ neighbor overlap
  with the full-context ranking (stability); (c) per-backbone break vs.\
  fix rates ($d_0$ vs.\ full, all ten backbones, sorted by \dofull).
  Context removal reshuffles neighbors and breaks far more than it
  fixes.}
  \label{fig:stratified}
\end{figure*}
\Cref{fig:stratified} accompanies the main paper's ``What Context
Removal Does'' subsection: panel (c) gives the per-backbone break vs.\
fix counts and panel (b) the neighbor-set overlap quoted there.

\section{Plain-background crossover table}
\label{sup:crossover}
Full similarity table for the main paper's ``plain-background
crossover'' analysis (\cref{tab:crossover}).

\begin{table}[t]
\centering\small
\setlength{\tabcolsep}{6pt}
\begin{tabular*}{\columnwidth}{@{\extracolsep{\fill}}llccc@{}}
\toprule
& & own cat. & plain bg. & cluttered bg. \\
\midrule
\multirow{2}{*}{all queries} & full & \textbf{.68} & .44 & .43 \\
                             & $d_0$ & \textbf{.50} & .38 & .36 \\
\midrule
\multirow{2}{*}{$d_0$ flips} & full & \textbf{.65} & .40 & .40 \\
                             & $d_0$ & .29 & \textbf{.31} & .28 \\
\bottomrule
\end{tabular*}
\caption{\textbf{Context removal flips similarity from the object to a
plain background} (DINOv2-G; cosine, the retrieval metric). Mean cosine
of the query descriptor to its own-category centroid vs.\ to plain- and
cluttered-background object centroids (background pools by image
background std, any category), at full context and $d_0$. With context
the descriptor is closest to its own category; for queries that break,
at $d_0$ it becomes closest to \emph{plain-background} objects---and more
so than to cluttered ones, so the driver is plainness, not a generic
collapse.}
\label{tab:crossover}
\end{table}

\section{Validity-check details}
\label{sup:validity}
Expanded methodology and mechanism analysis for the main paper's
``Stress tests'' paragraph (Tab.~3 of the main paper).

\paragraph{Fill and edge artifacts.}
The gray fill could create out-of-distribution statistics, and the hard
mask boundary could inject high-contrast edge tokens. Neither drives the
ranking: zero fill, a sharper RGB cut against natural images, preserves
it (Spearman $\rho{=}0.97$), and replacing the hard cut with an
alpha-blend feather---softening the boundary without revealing scene
content---leaves it unchanged ($\rho{=}1.00$, mean shift $+1.2$\,pp).
A Gaussian-blur fill preserves the local color and frequency statistics
of the surround rather than substituting an out-of-distribution gray
patch, and still leaves the ranking largely intact ($\rho{=}0.94$,
$p{<}10^{-4}$, mean shift $5.4$\,pp)---the largest shift among the
checks, driven mainly by the two strongest self-supervised backbones
(DINOv2-G, DINOv3-L), whose object-only retention rises by $9$--$12$\,pp
under blur relative to gray. A pixel-shuffle fill isolates why: it
conserves the background's color histogram \emph{exactly}---randomly
permuting the outside-mask pixels in place---but destroys the
low-frequency scene layout that a blur preserves. Under shuffle the
ranking again holds ($\rho{=}0.92$, mean shift $4.5$\,pp) and the blur
gains vanish: DINOv2-G and DINOv3-L fall back to at or slightly below
their gray-fill retention ($76.2\to62.8$ and $72.2\to55.2$). The
blur-specific rise is therefore residual low-frequency context leaking
through the fill---the encoder still reads coarse scene layout out of a
heavily blurred surround---rather than any artifact of the gray patch.
Under all three statistics-preserving fills every backbone's retention
still drops sharply from full image to object-only. The
real-background cut-paste condition (\cref{sup:cutpaste}) additionally
shows the phenomenon persists with no synthetic fill at all: retention
under a random real scene is $60$--$94\%$ across the panel, at least
as low as under gray fill for eight of ten backbones.

\paragraph{Mask-boundary robustness.}
KAP uses oracle masks, but real masks are imperfect. A per-mask random
rotation about the centroid ($\sigma_\theta{=}1^\circ$; median IoU
$0.96$ against the original mask over a $200$-mask sample) perturbs the
boundary while preserving object extent, and leaves the ranking
identical ($\rho{=}1.00$, mean shift $0.35$\,pp). The INSTRE and ILIAS
experiments close the loop with fully \emph{predicted} masks.

\section{Classifier-readout retention (all backbones)}
\label{sup:readout}
\Cref{tab:readout} gives the per-backbone classifier readout referenced
in the main paper (``A different readout agrees''). For each backbone we
train an $80$-way COCO-category linear probe on frozen full-image object
descriptors and measure classification retention
$\mathrm{acc}(d_0)/\mathrm{acc}(\text{full})$ under the same context-removal
intervention. The retention ranking tracks the retrieval \dofull{}
ranking (Spearman $\rho{=}0.87$), and the architecture dissociation
persists (ConvNeXt-L $62.2$ vs.\ ViT-L $57.7$), so context dependence is
a property of the descriptor, not of nearest-neighbor retrieval.

\begin{table}[t]
\centering\small
\setlength{\tabcolsep}{6pt}
\begin{tabular*}{\columnwidth}{@{\extracolsep{\fill}}lccc@{}}
\toprule
Backbone & acc.\ full & acc.\ $d_0$ & acc.\ $d_0$/full \\
\midrule
DINOv2-G/14    & 91.7 & 60.5 & 66 \\
ConvNeXt-L     & 80.0 & 49.7 & 62 \\
DINOv3-L       & 87.1 & 50.8 & 58 \\
ViT-L          & 82.1 & 47.4 & 58 \\
MAE ViT-L      & 58.3 & 23.6 & 41 \\
SigLIP2-512    & 90.0 & 53.4 & 59 \\
SigLIP2-L      & 87.4 & 48.8 & 56 \\
EVA-02-CLIP    & 83.5 & 42.7 & 51 \\
DINO ViT-S/8   & 64.7 & 24.6 & 38 \\
EVA-02-MIM     & 75.3 & 24.2 & 32 \\
\bottomrule
\end{tabular*}
\caption{\textbf{Classifier readout under context removal.} $80$-way
COCO-category linear-probe accuracy (\%) on full-image and object-only
($d_0$) descriptors, and their ratio acc.\ $d_0$/full---the
classification analogue of \dofull{}---for all ten backbones. The
ranking matches the retrieval \dofull{} ranking (Spearman
$\rho{=}0.87$).}
\label{tab:readout}
\end{table}

\section{Adapter training details}
\label{sup:lora}
The aperture-conditioned adapter (main paper, ``Reducing Context
Dependence'') is a rank-8 LoRA on the vision backbone's
\texttt{attn.qkv} projections (separate \texttt{q/k/v\_proj} on
EVA-02-MIM), trained with a teacher--student InfoNCE
objective on a single frozen backbone: the teacher view (LoRA disabled)
sees a wide aperture and the student view (LoRA enabled) a tight one,
both sampled per image; the student's global descriptor is pulled toward
the teacher's with in-batch negatives, and the teacher gradient is
stopped. The outside-mask fill is randomized across fill modes so the
adapter does not specialize to one. \Cref{tab:lorahp} lists the
hyperparameters; they are shared across the three adapted backbones.

\begin{table}[t]
\centering\small
\setlength{\tabcolsep}{8pt}
\begin{tabular*}{\columnwidth}{@{\extracolsep{\fill}}ll@{}}
\toprule
Hyperparameter & Value \\
\midrule
LoRA rank $r$            & $8$ \\
LoRA $\alpha$            & $16$ \\
LoRA dropout             & $0.1$ \\
Target modules           & \texttt{attn.qkv} \\
\quad on EVA-02-MIM      & \texttt{q/k/v\_proj} \\
Learning rate            & $2\times10^{-4}$ \\
Epochs                   & $1$ \\
Batch size               & $16$ \\
Training images          & $25{,}000$ (COCO \texttt{train2017}) \\
Image-sampling seed      & $43$$^\ast$ \\
Teacher aperture $d$     & $\sim\mathcal{U}(20,250)$\,px \\
Student aperture $d$     & $\sim\mathcal{U}(0,\min(200,d_{\text{teacher}}{-}10))$\,px \\
InfoNCE temperature      & $0.07$ \\
Outside-mask fills       & gray, zero, blur, noise \\
Trainable parameters     & $<\!0.4\%$ of backbone \\
\bottomrule
\end{tabular*}
\caption{\textbf{Aperture-conditioned LoRA hyperparameters.}
$^\ast$Disjointness from every evaluation query and gallery holds by
split: adapters are trained on
\texttt{train2017} and evaluated only on \texttt{val2017}, INSTRE, and
ILIAS.}
\label{tab:lorahp}
\end{table}

\section{Matched contrastive control (``SimCLR aug.'')}
\label{sup:simclr}
The no-aperture control (main paper, ``A matched control isolates the
aperture signal'') matches the aperture adapter in \emph{everything}
except the view construction: same rank-8 LoRA on the same attention projections,
same $25{,}000$ COCO \texttt{train2017} training images (identical
sampling: seed $43$, drawn after excluding the canonical
seed-$42$/$25$K subset, with the same per-image skip logic so the
effective image stream is identical), same teacher--student topology
(teacher = LoRA-disabled frozen backbone with stopped gradient,
student = LoRA-enabled), same one-directional InfoNCE with in-batch
negatives and temperature $0.07$, same learning rate
($2\times10^{-4}$), epochs ($1$), and batch size ($16$). The only
change: instead of a wide/tight aperture pair around a panoptic
segment, teacher and student each receive an independently augmented
view of the \emph{full frame} (no aperture, no mask):
\texttt{RandomResizedCrop} to the backbone's input size with scale
$\mathcal{U}(0.5,1.0)$ (bilinear), \texttt{RandomHorizontalFlip}, and
\texttt{ColorJitter}$(0.4,0.4,0.4,0.1)$.

We also trained a conventional \emph{symmetric} SimCLR variant (both
branches LoRA-enabled, no teacher, symmetric NT-Xent over the $2B$
in-batch views; same data, augmentations, and hyperparameters). It
shows the same qualitative pattern as the matched control: on ILIAS
full-image R@1 it reaches $69.4$ (EVA-02-MIM), $68.2$ (ViT-L), $80.2$
(SigLIP2-L) vs.\ the frozen $53.7/71.5/79.1$ and KAP's
$73.0/71.6/79.9$; on INSTRE (natural queries, per-backbone pooling) it
reaches $83.2/81.7/96.8$ R@1 vs.\ frozen $81.0/86.8/98.5$ and KAP's
$89.6/89.7/99.1$. Under either recipe, generic contrastive adaptation
helps only the weakest backbone and hurts the stronger ones, while the
aperture adapter improves or preserves all three.

\section{ILIAS positive-as-query protocol}
\label{sup:ilias}
ILIAS's provided queries are already context-removed (clean or
bare-surface backgrounds), so they cannot serve as a masking
counterfactual. We quantified this directly. A photographed backdrop
always carries sensor noise and a lighting gradient, so its pixels
scatter by several units even when it looks uniform; a digitally filled
background stays \emph{exactly} constant (a flat region survives JPEG's
DCT near-losslessly). We therefore count a query as a hard-segmented
cutout when (i) the modal border color occupies $\ge20\%$ of the frame at
exact equality and (ii) that same color also covers $\ge5\%$ of the
pixels \emph{inside} the annotated box---i.e.\ the fill wraps into the
object's concavities, which letterbox bars and studio sweeps do not.
By this test $211/1{,}232$ queries ($17.1\%$) are cutouts, spanning
$192/1{,}000$ instances, with fills near-white ($124$), near-black ($65$)
or other ($22$); the identical test on the natural positives, which pass
through the same capture and JPEG pipeline, returns $27/4{,}715$
($0.6\%$), confirming the signature is digital fill rather than a
compression or annotation artifact. We instead build, for each of the $938$ instances with at least two annotated positives, a
query from one randomly chosen \emph{cluttered} positive (fixed seed
$0$) and retrieve against the $4{,}653$ natural positives with the
query's own image excluded; same-instance images are the relevants.
ILIAS \texttt{bbox.json} boxes are COCO-style $[x,y,w,h]$ (verified:
$0/4{,}715$ boxes invalid under this reading vs.\ $60\%$ invalid read as
$[x_0,y_0,x_1,y_1]$). Query masks are SAM3 box-prompted segmentations of
the annotated box; a mask is accepted if its IoU with the box region is
$\ge0.2$, else the box itself is used as the mask (fallback). Context
removal gray-fills outside the (undilated) mask and the descriptor is
always pooled over the original mask, exactly as in the COCO protocol;
features come from the same extractor code path as all other
experiments.

\section{Real-background cut-paste counterfactual}
\label{sup:cutpaste}
The natural-background control (main paper, ``Fill and edge
artifacts'') requires no synthetic fill: for each INSTRE query, all
pixels outside the SAM3 object mask are replaced by a randomly drawn COCO
scene, bilinearly resized to the query image's resolution. Queries are
pooled over the object mask; masks with IoU $<0.2$ against the annotated
box are dropped ($1{,}208/1{,}250$ queries kept). The gallery is the
natural $27{,}293$-image INSTRE gallery, identical across the natural /
gray-fill / cut-paste conditions; confidence intervals are query
bootstraps.

\section{Pooling operators}
\label{sup:pooling}
The pooling sweep (main paper, ``Pooling operator'') compares three
parameter-free operators over the object's feature tokens: \emph{mean}
(the paper's default), \emph{max} (coordinate-wise maximum), and
\emph{attentive} (tokens weighted by a softmax over their feature
$\ell_2$ norms---a salience weighting with no learned parameters).
Query and gallery always use the same operator, and each descriptor is
$\ell_2$-normalized.

\section{Seeds and reproducibility}
\label{sup:seeds}
Query sampling for all COCO sweeps is balanced per category
(40/category, 80 categories) with seed $42$; adapter and control
training use seed $42$ with image-sampling seed $43$. Disjointness
holds by split: adapters are trained on \texttt{train2017} images and
evaluated only on COCO \texttt{val2017}, INSTRE, and ILIAS, so no
adapter-training image occurs in any gallery used to evaluate an
adapted model; ILIAS positive-as-query sampling uses seed $0$;
bootstrap confidence intervals use $10^4$ paired query resamples
throughout. Code, adapter checkpoints, and the exact query lists are available at
\url{https://github.com/bmeivar/kap}.

\end{document}